\documentclass[twoside]{article}

\usepackage[accepted]{aistats2021}
\usepackage{amsfonts,amsthm}
\usepackage{graphicx}
\graphicspath{{images/}}
\usepackage{subcaption}
\usepackage{enumerate}
\usepackage{soul}
\usepackage{color}
\usepackage{hyperref}
\usepackage[ruled,vlined]{algorithm2e}
\usepackage{doi}
\usepackage{mathtools}

\newcommand{\x}{{\bf x}}
\newcommand{\code}[1]{\text{{\fontfamily{lmtt}\selectfont #1}}}

\newtheorem{thm}{Theorem}

\newtheorem{prop}{Proposition}
\newtheorem{exmp}{Example}
\newtheorem{assump}{Assumption}

\theoremstyle{remark}
\newtheorem{remark}{Remark}

\SetKw{Continue}{continue}

\makeatletter
\renewcommand\@biblabel{}
\makeatother

\hypersetup{colorlinks=true,allcolors=blue}

\usepackage[round]{natbib}

\begin{document}



\twocolumn[

\aistatstitle{Product Manifold Learning}

\aistatsauthor{Sharon Zhang \And Amit Moscovich \And Amit Singer}

\aistatsaddress{ Princeton University \And Princeton University \And Princeton University} ]

\begin{abstract}
We consider problems of dimensionality reduction and learning data representations for continuous spaces with two or more independent degrees of freedom. Such problems occur, for example, when observing shapes with several components that move independently. Mathematically, if the parameter space of each continuous independent motion is a manifold, then their combination is known as a product manifold. In this paper, we present a new paradigm for non-linear independent component analysis called manifold factorization. Our factorization algorithm is based on spectral graph methods for manifold learning and the separability of the Laplacian operator on product spaces. Recovering the factors of a manifold yields meaningful lower-dimensional representations and provides a new way to focus on particular aspects of the data space while ignoring others. We demonstrate the potential use of our method for an important and challenging problem in structural biology: mapping the motions of proteins and other large molecules using cryo-electron microscopy datasets. 
\end{abstract}

\section{Introduction}
Consider a data-generating process $F$ which maps vectors of latent (unobserved) variables to observations,
\begin{align}
    (\theta_1,\ldots, \theta_m) \xmapsto{\ F\ } \x \in \mathbb{R}^D.
\end{align}
Our focus is on the case where the latent variables are low-dimensional and the observations are  high-dimensional vectors.
As an illustrative example, consider an industrial articulated robot where the latent variables $\theta_1, \ldots, \theta_m$ correspond to the angles of its $m$ rotary joints, and the observed vector $\x$ is a set of measurements recorded by an external monitoring system. When $F$
% the mapping $(\theta_1,\ldots, \theta_m) \mapsto \x$ 
is deterministic (i.e. without noise), full-rank, smooth and injective, then the space of observations is a submanifold of $\mathbb{R}^D$ \citep[Theorem 4.14]{Lee2012}.
This observation underlies the field of \emph{manifold learning} \citep{TenenbaumDesilvaLangford2000,BelkinNiyogi2003,CoifmanLafon2006,TalmonCohenGannotCoifman2013}.
A key problem in this field is dimensionality reduction, where a set of high-dimensional observations $\x_1, \ldots, \x_n \in \mathbb{R}^D$ are mapped to a low-dimensional space, preferably one whose dimension is not much larger than the dimension of the latent space.

In this work, we explore the idea of manifold learning when the latent space is a \emph{product manifold}.
If each latent variable $\theta_i$ lies on a manifold $\mathcal M_i$ of dimension $d_i$ then the latent space is the cartesian product
\begin{align} \label{eq:manifold_factorization}
    \mathcal{M} = \mathcal{M}_1 \times \cdots \times \mathcal{M}_m
\end{align}
which is a manifold of dimension $d = d_1 + \cdots + d_m$.
We  refer to the decomposition \eqref{eq:manifold_factorization} as a \textit{manifold factorization} of $\mathcal{M}$ and to each of the manifolds $\mathcal{M}_i$ as \textit{factors}.
Learning in the latent space is achieved by a data-dependent manifold factorization algorithm whose input is the output of spectral graph-based methods for manifold learning \citep{BelkinNiyogi2003,CoifmanLafon2006}.
These algorithms output a set of $N$ Laplacian eigenvectors which approximate the Laplaceian eigenfunctions of the manifold.
On a product manifold the eigenfunctions are nothing but products of eigenfunctions on the factor manifolds.
Hence, we can try to find the eigenvectors of the factor manifolds by seeking two small sets of vectors whose cartesian elementwise products produce all of the $N$ eigenvectors.

Instead of having each observation map to a point on a high-dimensional manifold, our approach enables each observation to be mapped to several points on separate low-dimensional spaces.
Since each factor may encode a different aspect of the data, this opens the door to exciting new methods for data visualization and representation.
For example, one can use it to visualize observations which come from a high-dimensional latent spaces as a set of points on $m$ low-dimensional manifolds that correspond to the factors of $\mathcal M_1, \ldots, \mathcal M_m$, then determine which aspects of the data space correspond to what factor, and ultimately focus only on the factors of interest.

\subsection{Related Work}

In this paper, we build on spectral embedding methods for dimensionality reduction and data representation \citep{BelkinNiyogi2003,CoifmanLafon2006}.
These methods construct a data-dependent graph Laplacian which approximates the Laplacian operator on the data manifold.
Furthermore, the eigenvectors of the graph Laplacian converge to the eigenfunctions of the manifold Laplacian.
By mapping the observations to their eigenvector coordinates one can obtain a low-dimensional embedding of the data space \citep{Bates2014}. 

One work that is closely related to ours is the spectral method of \citep{Singer2006a} for linear independent component analysis (ICA), as it is also based on the separability of the Laplacian on a product space.
This was later extended to a method for non-linear ICA that assumes that the latent space has a product structure and that the Jacobian of the mapping $(\theta_1,\ldots, \theta_m) \mapsto \x$ is either known or can be estimated in some way \citep{SingerCoifman2008}.
Recently, \citep{Rodola2019} applications of spectral representations on product manifolds to shape analysis problems such as finding correspondences.

In the neural-network literature, learning \emph{disentangled representations} has drawn interest in recent years \citep{ReedSohnZhangLee2014,LocatelloEtal2019,Tran2017,Siddharth2017, Kim2018}.
Unlike our work, which makes strict assumptions on the data manifold,
the disentanglement literature seems to treat broader classes of data-generating processes and does not have a formal notion of what constitutes a disentangled representation.

\section{Theoretical background}\label{theory}

We begin with a brief review of some useful properties of the Laplace operator, and then examine how the Helmoltz equation separates over product domains in $\mathbb{R}^d$. Lastly, we go over results that relate the graph Laplacian to the continuous Laplacian, and introduce the connection to diffusion maps.

\subsection{Properties of the Laplace operator}

Let $f$ be a real-valued, twice-differentiable function and $\mathcal{M}$ a compact $d$-dimensional manifold throughout this section.
Let $\nabla=\nabla_{\mathcal M}$ be the gradient and $\Delta = \Delta_{\mathcal M} = \nabla_{\mathcal M} \cdot \nabla_{\mathcal
M}$ the manifold Laplacian (or Laplace-Beltrami operator).
We consider the solutions  of the Helmholtz equation with Neumann boundary conditions
\begin{equation}\label{helmholtz}
    -\Delta f(\mathbf{x}) = \lambda f(\mathbf{x}), \quad \forall \mathbf{x} \in \mathcal{M}
\end{equation}
\begin{equation}\label{neumann-bc}
    \nabla f(\mathbf{x}) \cdot \nu(\mathbf{x}) = 0, \quad \forall \mathbf{x} \in \partial \mathcal{M}
\end{equation}
where $\nu$ is a normal unit vector to the boundary.
The functions $f$ are known as the \textit{Neumann eigenfunctions}, and together with the corresponding eigenvalues $\lambda$ they comprise the spectrum of the Laplacian operator on $\mathcal{M}$. In this paper, we will only consider the particular Neumann boundary condition specified in (\ref{neumann-bc}), so we refer to the Neumann eigenfunctions simply as the eigenfunctions. A well-known property of the eigenfunctions is the following.

\begin{thm}\label{laplace-props}
Let $\{f_k\}_{k=1}^\infty$ be the Neumann eigenfunctions, and $\{\lambda_k\}_{k=1}^\infty$ be the corresponding eigenvalues of a compact domain $\Omega \subset \mathbb{R}^d$. Then 
\begin{enumerate}[(i)]
    \item $0 = \lambda_1 \le \lambda_2 \le \lambda_3 \le \cdots$ and $\lambda_k \rightarrow \infty$ as $k \rightarrow \infty$.
    \item The eigenfunctions $\{f_m\}_{m=1}^\infty$ form a complete orthonormal basis for $L^2(\Omega)$.
\end{enumerate}
\end{thm}

For a complete proof and discussion on generalizations, see \citep{Folland1976}. In particular, the second result of Theorem \ref{laplace-props} tells us that we can express any square-integrable function over the data manifold in terms of the eigenfunctions. Analogs of these results also hold for more general Riemmannian manifolds.

\subsection{The Laplacian over product spaces} \label{sec:laplacian_product}

As we shift to the case when our data comes from a product manifold $\mathcal{M}$, we can take advantage of some key separability properties of the Laplacian. For simplicity, we treat the case when $\mathcal{M} = \mathcal{M}_1 \times \mathcal{M}_2$.

\begin{prop}\label{laplace-product}
Let $f_1 : \mathcal{M}_1 \rightarrow \mathbb{R}$ and $f_2 : \mathcal{M}_2 \rightarrow \mathbb{R}$ be twice-differentiable functions such that $\Delta_{\mathcal{M}_1}f_1 = \lambda_1 f_1$ and $\Delta_{\mathcal{M}_2}f_2 = \lambda_2 f_2$. Take $\pi_i : \mathcal{M} \rightarrow \mathcal{M}_i$ to be the projection of $\mathcal{M}$ onto $\mathcal{M}_i$ for $i = 1,2$. We can then define the natural extension of $f_i$ to $\mathcal{M}$ via $g_i = f_i \circ \pi_i$. It follows that
\(
    \Delta_\mathcal{M}(g_1g_2) = (\lambda_1 + \lambda_2)g_1g_2.
\)
\end{prop}
%\begin{proof}
%Using the product rule, 
%\begin{align}
%    \notag \Delta_{\mathcal{M}}&(g_1g_2) = g_1 \Delta_{\mathcal{M}}g_2 + g_2 \Delta_{\mathcal{M}}g_1 - 2 \langle \nabla_\mathcal{M} g_1, \nabla_\mathcal{M} g_2 \rangle \\
%    \notag &= -g_1 \nabla \cdot ((\mathbf{J}_{\pi_2})^T \nabla f_2) - g_2 \nabla \cdot ((\mathbf{J}_{\pi_1})^T \nabla f_1) \\
%    \notag &= g_1 (\mathbf{J}_{\pi_2})^T \Delta_{\mathcal{M}_2}f_2 + g_1 (\mathbf{J}_{\pi_1})^T \Delta_{\mathcal{M}_1}f_1 \\
%    \notag &= (\lambda_1 + \lambda_2)g_1g_2.
%\end{align}
%Here, $\mathbf{J}_{\pi_i}$ denotes the Jacobian matrix of the projection $\pi_i$.
%\end{proof}
For a proof, see for example \citep[Section 4.6]{Canzani2013}.
Proposition \ref{laplace-product} tells us that the product of the eigenfunctions of $\mathcal{M}_1$ and $\mathcal{M}_2$ are eigenfunctions of $\mathcal{M}$ and that the corresponding eigenvalue is equal to the sum of the eigenvalues in $\mathcal{M}_1$ and $\mathcal{M}_2$. Note that this result easily extends to the general product of $m > 2$ Riemannian manifolds.

Thus far, we can conclude that products of eigenfunctions of manifold factors are eigenfunctions of the product manifold. The converse is actually also true, and this can be shown by an application of the Stone-Weierstrass theorem \citep{Canzani2013}. Hence, the eigenfunctions of $\mathcal{M} = \mathcal{M}_1\times\mathcal{M}_2$ are precisely
\begin{equation}
    \{f_ig_j \mid \Delta_{\mathcal{M}_1}f_i = \lambda_if_i, \; \Delta_{\mathcal{M}_2}g_j = \mu_jg_j \}
\end{equation}
with corresponding eigenvalues $\{\lambda_i+\mu_j\}_{i,j}$.
For the $m > 2$ case, the eigenfunctions of $\mathcal{M}$ are of the form
\begin{equation}
    f^{(k_1, \ldots, k_m)} = \prod_{i=1}^m f^{(i)}_{k_i},
\end{equation}
with corresponding eigenvalues
\begin{equation}
    \lambda^{(k_1, \ldots, k_m)} = \sum_{i=1}^m \lambda^{(i)}_{k_i}.
\end{equation}
Note that since $\lambda^{(i)}_0 = 0$ and $f^{(i)}_0 = \big(\sqrt{\text{vol}(\mathcal{M}_i)}\big)^{-1}$ for all $i = 1, \ldots, m$, we must have $\lambda^{(\mathbf{0})} = 0$ and $f^{(\mathbf{0})} = \prod_i\big(\sqrt{\text{vol}(\mathcal{M}_i)}\big)^{-1}$ as well. This is consistent with (\ref{laplace-props}).

\begin{exmp}[2D rectangle]
    Consider the domain
    \begin{align}
        \mathcal{M} = [0, a] \times [0, b] \subset \mathbb{R}^2.
    \end{align}
    The Laplacian eigenfunctions with Neumann boundary conditions are
    \begin{equation}\label{eq4}
        f_{m,n}(x,y) = \tfrac{1}{2}\cos\left(\tfrac{m\pi}{a}x\right)\cos\left(\tfrac{n\pi}{b}y\right),
    \end{equation}
    with eigenvalues
    \begin{equation}\label{eq5}
\lambda_{m,n} = \lambda_m + \lambda_n = \pi^2\left(\tfrac{m^2}{a^2} + \tfrac{n^2}{b^2}\right).
\end{equation}
The eigenfunctions are precisely the product of the eigenfunctions on the closed intervals $[0, a]$ and $[0, b]$, which are $\cos(\frac{m\pi}{a}x)$ and $\cos(\frac{n\pi}{b}y)$. The eigenvalues of the product space are also the sums of the corresponding eigenvalues of the intervals.
\end{exmp}
For the Laplacian eigenfunctions and eigenvalues on other domains, see the review by \cite{GrebenkovNguyen2013}.
 
\subsection{Graph Laplacians and diffusion maps}
The results of the previous sections hold for the continuous Laplacian, but in practice we only have limited observable samples $\mathbf{x}_1, \ldots \mathbf{x}_n \in \mathbb{R}^D$. Thus, we must turn to the discrete analog of the Laplace operator, the graph Laplacian. To start, define the fully-connected graph $(G,E,V)$ whose edge weight matrix $W \in \mathbb{R}^{n \times n}$ is described by 

\begin{equation}
W_{ij} = \exp \left(-\|x_i-x_j\|^2/\epsilon \right).
\end{equation}

We then compute the random walk matrix $A = D^{-1}W$, where $D$ is the diagonal degree matrix,
\begin{equation}\label{degree_matrix}
D_{ii} = \sum_{j=1}^n W_{ij}.
\end{equation}
The matrix $A$ can be viewed as a transition probability matrix, and the $i$th entry of the $j$th column of $A^t$ describes the probability cloud of a random walk on $G$ ending up in position $j$ at time $t$, given a starting position at $i$. The diffusion map \citep{CoifmanLafon2006, BelkinNiyogi2003} is a mapping $\phi : V \rightarrow \mathbb{R}^{n-1}$ given by
\[
\phi_t(v_i) = [\lambda^t_2 \phi_2(i), \lambda^t_3 \phi_3(i), \ldots, \lambda^t_n \phi_n(i)]
\]
Here, $\phi_j$ are the right eigenvectors of $A$ and $\phi_j(i)$ denotes the $i$th coordinate of $\phi_j$. The coordinates of the embedding are known as diffusion coordinates. It can be shown that the diffusion distance, which is defined as the Euclidean distance in the diffusion coordinates, corresponds to the difference between different probability clouds \citep{CoifmanLafon2006}. Alternatively, if we view the observed data as samples from a smooth signal on the manifold, then the eigenvectors can also be interpreted as a Fourier basis for these signals \citep{CoifmanLafon2006, LeeIzbicki2016}.

The standard graph Laplacian is defined as $L = D - W$. We can also define a normalized version called the random walk Laplacian, which is $L_{rw} = D^{-1}L = I - A$. The random walk Laplacian has several properties that are discussed in the next section which make it the most suitable approximation. Computing the eigenvectors and eigenvalues of $L_{rw}$ directly is computationally inefficient, but we can compute them indirectly via the spectrum of the symmetric Laplacian, $L_{sym} = D^{-1/2}LD^{-1/2}$. The spectrum of $L_{rw}$ can then be derived directly from the spectrum of $L_{sym}$ \citep{VonLuxburg2007}. A condensed version of this procedure can also be found in the thesis of \cite{Lafon2004}. 

\subsection{Limit of the graph Laplacian} \label{sec:graph_laplacian_convergence}

The graph Laplacian constructed from an i.i.d. sample of points on a manifold converges to a linear differential operator.
For the standard kernel-based construction of the graph and when the samples are drawn uniformly the limiting operator is the Laplace-Beltrami operator $\Delta_{\mathcal M}$ \citep{HeinAudibertLuxburg2005,Singer2006b,BelkinNiyogi2008}.
If the data instead has a non-uniform density
$p(\x)$, the graph Laplacian converges to the Fokker-Planck operator on the manifold, which has an additional drift term given by the log-density $U(x) = -2\log p(x)$ \citep{NadlerLafonCoifmanKevrekidis2005, CoifmanLafon2006},
\begin{equation}\label{fokker-placnk}
    \mathcal{L}f = \Delta_{\mathcal M} f - \nabla U \cdot \nabla f.
\end{equation}
Alternative graph constructions, such as the k-nearest-neighbor graph also converge to a Fokker-Planck operator \citep{TingHuangJordan2010}.

In addition to the pointwise convergence, several works have proved the spectral convergence of the graph Laplacian.
i.e. the convergence of the eigenvalues and eigenvectors of graph Laplacians \citep{VonLuxburgBelkinBousquet2008,RosascoBelkinDevito2010,TrillosSlepcev2018,TrillosGerlachHeinSlepcev2020}.

\section{Method}\label{method}

As discussed in Section \ref{sec:laplacian_product}, each eigenfunction $\varphi_k$ on the product space $\mathcal{M} = \mathcal{M}_1 \times \mathcal{M}_2$ is the product of an eigenfunction $\varphi_i$ on $\mathcal M_1$ and an eigenfunction $\varphi_j$ on $\mathcal M_2$.
A corollary of the convergence of the graph Laplacian (see Section~\ref{sec:graph_laplacian_convergence}) is the following: given a large sample of points on $\mathcal M$ (or any isometric embedding of it), if we construct the graph Laplacian from the sample then each Laplacian eigenvector $\varphi_k$ should be approximately equal (up to normalization) to an elementwise product $\varphi_i \varphi_j$ where $\varphi_i$ and $\varphi_j$ approximate eigenfunctions on $\mathcal M_1$ and $\mathcal M_2$ respectively.
If either $\varphi_i$ or $\varphi_j$ correspond to an eigenfunction with eigenvalue $\lambda=0$ we shall say $\varphi_k$ is a \textit{factor} eigenvector.
Otherwise, we call $\varphi_k \approx \varphi_i \varphi_j$ a \textit{product} eigenvector.
Note that for data sampled uniformly from a connected manifold, the  only eigenfunctions with $\lambda=0$ are the constant functions.

The main idea behind our method is to first differentiate between the factor eigenvectors and the product eigenvectors and then to divide the factor eigenvectors into two sets that correspond to the eigenspace of $\mathcal M_1$ and $\mathcal M_2$. This is based on the observation that if $\varphi_k \approx \varphi_i \varphi_j$ then this is a hint that $\varphi_i$ and $\varphi_j$ belong to different factor manifolds.
Mathematically, our method is based on two assumptions:
\begin{assump} \label{assump:isometric}
    The mapping $(\theta_1,\ldots, \theta_m) \mapsto \x$ is an isometric embedding into $\mathbb{R}^D$.
\end{assump}

\begin{remark}
    Embeddings of Riemannian manifolds into Euclidean space exists for any Riemannian manifold \citep{Nash1954,Kuiper1955}.
    As isometric embeddings preserve the Riemannan metric, the spectral properties are not affected by the specific embedding.
\end{remark}

\begin{assump}
    The latent variables $\theta_1,\ldots, \theta_m$ are drawn independently of each other.
\end{assump}
\begin{remark}
    This assumption may be relaxed by using the diffusion maps normalization for the graph Laplacian \citep{CoifmanLafon2006}.
In contrast to the random-walk graph Laplacian, which converges to a density-dependent Fokker-Planck operator, with the diffusion
maps normalization the graph Laplacian converges to the Laplace-Beltrami operator which does not depend on the density.
\end{remark}

We now present an algorithm for factoring the set of eigenvectors of the graph Laplacian.

\subsection{Factorizing product eigenvectors}

First, we find for each eigenvector $\varphi_k$ the pair $\varphi_i, \varphi_j$ whose elementwise product is closest.
This ``closeness" is measured by the absolute cosine similarity,
\begin{align}
    \text{S}(\varphi_k,\, \varphi_i \varphi_j )
    :=
    \frac{ \big| \langle \varphi_k,\, \varphi_i \varphi_j \rangle \big|}{ \| \varphi_k \| \| \varphi_i \varphi_j\|}
    \in
    [0,1].
\end{align}
We take the absolute value of the dot product because of the sign ambiguity of the computed eigenvectors.
For each $k$, we find the combination $(i, j)$ with the highest similarity. 

To avoid computing the element-wise product for all $k(k-1)/2$ combinations, we skip triplets $(i, j, k)$ for which $|\lambda_i + \lambda_j - \lambda_k| > \delta$ for some certain eigenvalue threshold $\delta > 0$, which is specified as a parameter. This ``eigenvalue criterion" is based on the results of Proposition~\ref{laplace-product}. In terms of the quality of the match, the eigenvalue criterion may only act as a coarse filter. Thus, we additionally filter out triplets whose highest similarity is less than some threshold $\gamma$, which we call the ``similarity criterion." In total, we store at most $N$ combinations, one for every $1 \le k \le N$. This process is summarized in Algorithm \ref{alg1}.

\begin{algorithm}
\SetAlgoLined
\KwData{Eigenvectors $\{\varphi_1, \ldots, \varphi_N\}$ of $L_{rw}$.}
\KwResult{List of triplets $(i,j,k)$ where $\varphi_k \approx \varphi_i \varphi_j$ and their corresponding similarity scores.}

    \For{$k \leftarrow 1 \ldots N$}{
        \code{maxS} $\gets 0$\;
        \For{$i,j<k$}{
            \If{$|\lambda_i+ \lambda_j - \lambda_k| < \delta$ \code{and} \\ \text{S}$(\varphi_k, \varphi_i \varphi_j) >$ \code{maxS}}{
                \code{maxS} $\gets$ S$(\varphi_k,\, \varphi_i \varphi_j)$\;
                $i_{\max} \gets i$\;
                $j_{\max} \gets j$\;
            }
    }
    \If{\code{maxS} $> \gamma$}{
        add $(i_{\max},\, j_{\max},\, k)$ to \code{triplets}\;
    }
}
\caption{Indentification of individual factors}
\label{alg1}
\end{algorithm}

\begin{figure*}[ht]
    \centering
    \includegraphics[width=0.31\linewidth]{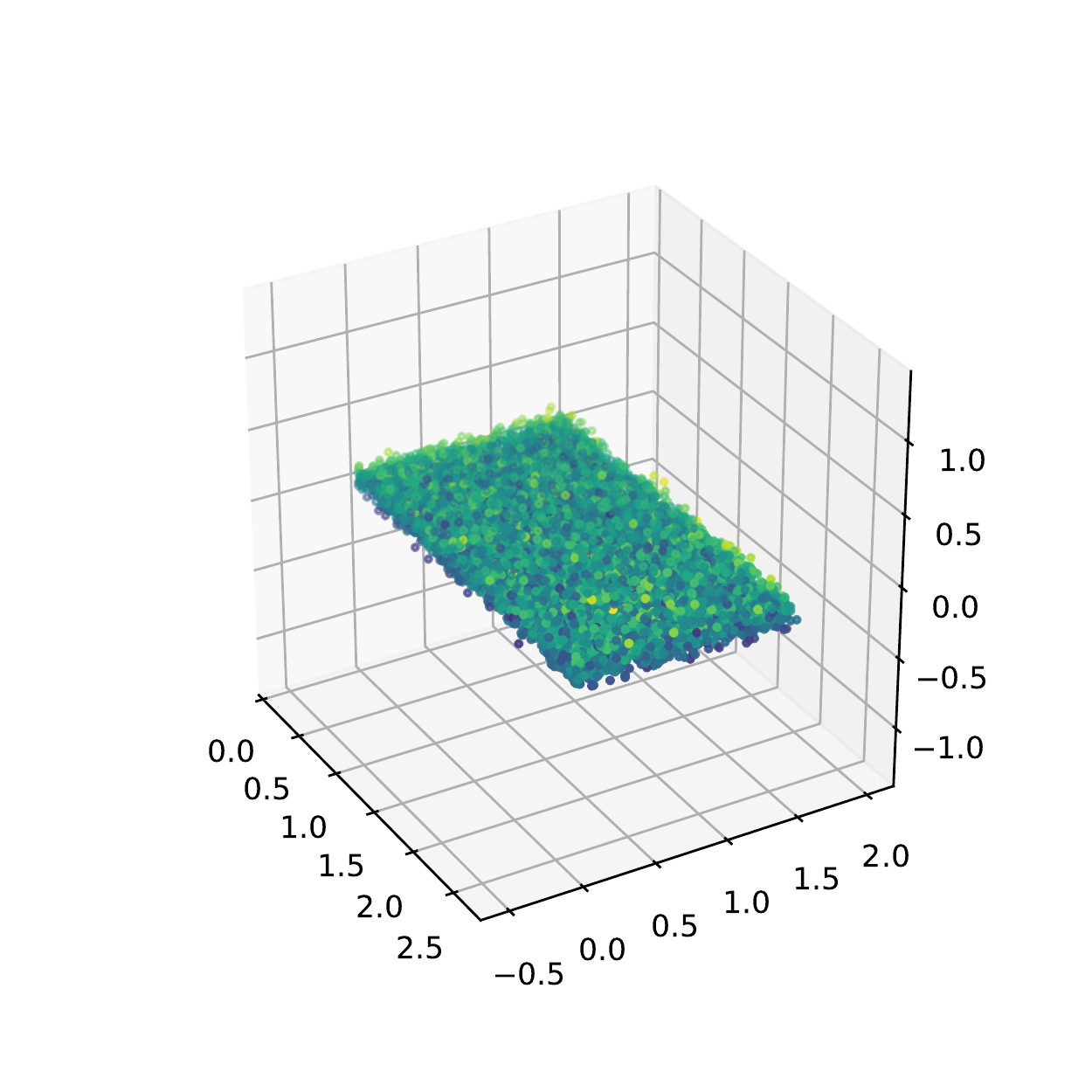}
    \hspace{1cm}
    \includegraphics[width=0.3\linewidth]{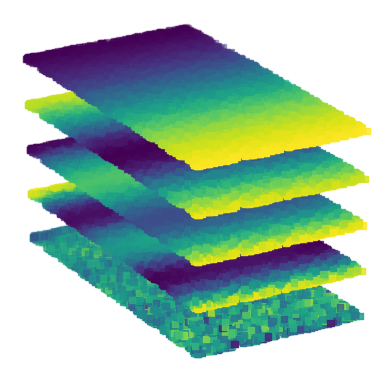}
    \includegraphics[width=0.3\linewidth]{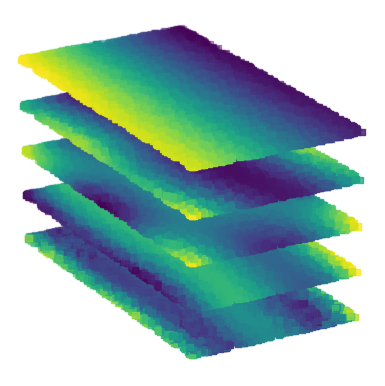}
    \caption{\textbf{Rectangle with noise.} The algorithm was run on 10,000 samples with $\delta = 0.5$, $\gamma = 0.85$, $K = 0$. For clarity, the eigenvectors shown are plotted on the non-noisy ground truth $x$ and $y$ coordinate plane. \textbf{Left.} The original data, with depth of noise indicated by color. \textbf{Middle.} The first five factor eigenvectors associated with the line $[0, 1 + \sqrt{\pi}]$ as determined by the algorithm. \textbf{Right.} The first five factor eigenvectors associated with the line $[0, 1.5]$ as determined by the algorithm.}
    \label{fig:rectangle3d_manifolds}
\end{figure*}

\begin{figure*}[ht]
    \centering
    \includegraphics[width=0.31\linewidth]{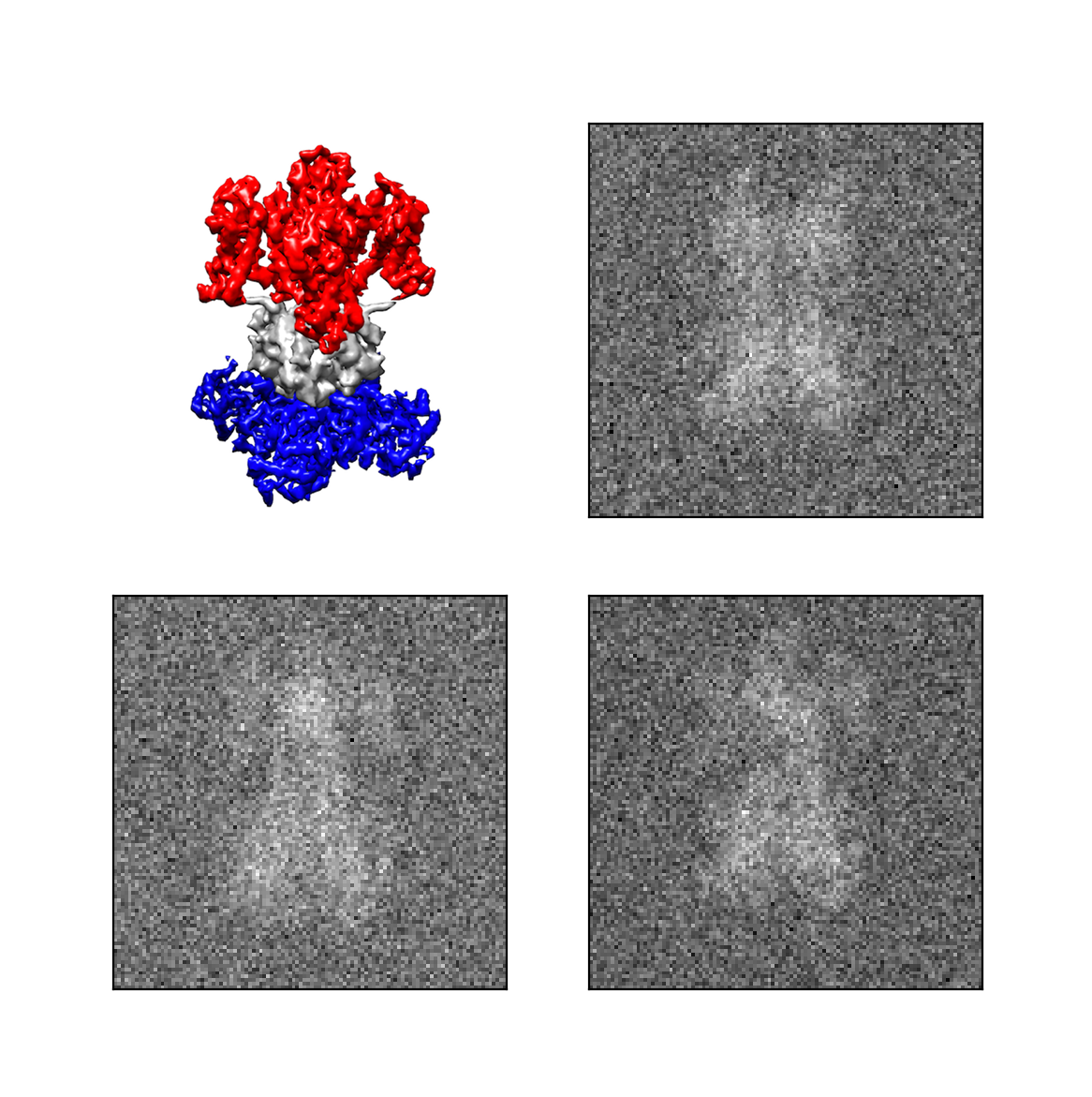}
    \hspace{0.75cm}
    \includegraphics[width=0.3\linewidth]{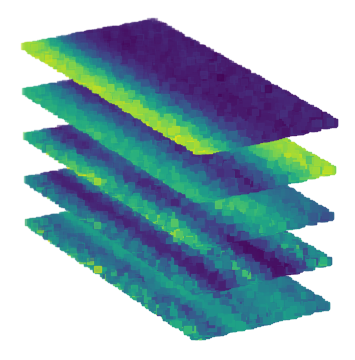}
    \includegraphics[width=0.3\linewidth]{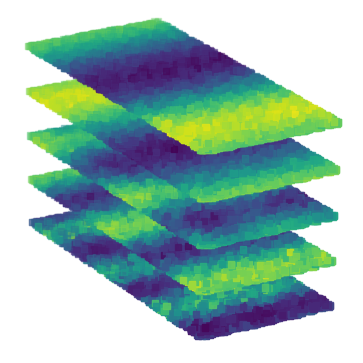}
    \caption{\textbf{Two-factor noisy cryo-EM data.} The algorithm was run on 10,000 images with the parameters $\delta = 1.0$, $\gamma = 0.80$. For clarity, the eigenvectors shown are plotted on the ground-truth values of the rotation angles $[0, 360]$ and the $x$-stretch $[-20, 20]$. \textbf{Left.} A synthetic cryo-EM image of the potassium ion channel with two independently deformable subunits is shown in the top-left. The red portion rotates randomly around the $z$-axis, and the blue portion stretches independently along the $x$-axis. Simulated cryo-EM images are shown in the other squares. \textbf{Middle.} The factor eigenvectors associated with $x$-stretch. \textbf{Right.} The factor eigenvectors associated with the rotation.
    }
    \label{fig:cryo_x-theta_noisy_manifolds}
\end{figure*}

The result of the algorithm is a list of index triplets $(i,j,k)$, indicating that the best factorization we could find for $\varphi_k$ is $\varphi_i \varphi_j$.
Next, we use the triplets to divide the factor eigenvectors into two sets that correspond to the manifolds $\mathcal M_1$ and $\mathcal M_2$.

\subsection{Assigning the factor eigenvectors to factor manifolds}\label{separation}

We can use the list of triplets to split the candidate factor eigenvectors into two separate subsets using a \textsc{Max-Cut} algorithm.

To do so, we store the computed similarity scores in a $T \times T$ \textit{separability matrix} $C$, where $T$ is the number of eigenvectors which appear as a factor eigenvector in any triplet. Each entry of $C$ is defined by
\begin{align}
    C_{i,j} = \text{S}(\varphi_k,\, \varphi_i \varphi_j),
\end{align}
and note that $C$ is upper-triangular with zero diagonal. Here, the similarity score is repurposed as a representation of the factorization assignments between each pair of factor eigenvectors in a triplet. The separation of two eigenvectors into eigenspaces of distinct manifold factors should be prioritized depending on the similarity of the triplet in which they appear. This naturally lends to a graph-cut approach to determine the ideal separation.

We use the symmetrized separability matrix, $C + C^T$, as the input to our \textsc{Max-Cut} algorithm to obtain two groups of factor eigenvectors. Since the original \textsc{Max-Cut} problem is NP-hard, we use the semi-definite program (SDP) relaxation described by \citep{Goemans1995}.

\subsection{Implementation details and runtime}

The entire algorithm was implemented in Python, and the \textsc{Max-Cut} SDP was implemented using CVXPY \citep{siamond2016cvxpy}. The first part of the algorithm involves computing the diffusion map of the data. The runtime for this step is dependent on the Scikit-learn randomized SVD algorithm \citep{scikitlearn2011, Martinsson2011}. In the subsequent step, we must find the best triplet for each of $k = 1, \ldots, N$ out of all $k(k-1)/2$ triplets. Taking into account the element-wise vector multiplication at each step, this takes $O(nN^3)$ in the worst case. In practice, however, the eigenvalue criterion eliminates the multiplication computation for the vast majority of triplets. The last part of the algorithm consists of iterating through the list of triplets to form the dissimilarity matrix $C$ and then using a \textsc{Max-Cut} SDP solver. Table \ref{runtime-table} gives a breakdown of runtimes for each of these parts on $10,000$ samples from a rectangle with three-dimensional noise.

\begin{table}[ht]
\begin{center}
\caption{Running times (in seconds) for each part of the algorithm on a 2017 2.9 GHz Quad-Core Intel Core i7 processor. The experiment was run for various $N$ on $10,000$ samples from a rectangle $\mathcal{M} = [0, 1 + \sqrt{\pi}] \times [0, 1.5]$ with Gaussian noise $\mathcal{N}(0, 0.1)$ in the $z$-direction, where $N$ is the total number of computed eigenvectors. The algorithm parameters were set to $\delta = 2.0$, $\gamma = 0.75$. For each $N$, we take the average runtime over five trials. The third column refers to Algorithm~\ref{alg1} and the fourth column refers to the construction of the separability matrix and the \textsc{Max-Cut} SDP described in Section~\ref{separation}.}
\begin{tabular}{cccc}
\textbf{$N$}  & \textbf{Diffusion map} & \textbf{Alg. 1} & \textbf{Alg. 2} \\
\hline
50  &  5.55&   1.24 &  0.03 \\
100 &  5.98 &   7.54 & 0.026 \\
200 &  8.30 &  43.19 &  0.03 \\
400 & 12.49 & 251.68 & 0.044 \\
\hline
\end{tabular}
\label{runtime-table}
\end{center}
\end{table}

\section{Simulations}\label{simulations}

We tested the algorithm on two different types of data: uniform samples from a noisy rectangle and synthetic cryo-EM data.

\subsection{2D noisy rectangle}
As a baseline, we ran the algorithm on data sampled from a rectangle that lies in $\mathbb{R}^3$. The data was generated by taking random samples from a  product of lines $\ell_1 = [0, \sqrt{\pi} + 1]$ and $\ell_2 = [0, 1.5]$, and then adding Gaussian noise in the $z$-direction. We ran two different trials on the rectangle data: first, using 10,000 samples with Gaussian noise applied in the $z$-direction; second, repeating the first trial except with additional Gaussian noise added in the $x$ and $y$ directions.

For each trial, we use $\gamma = 0.85$ and $\delta = 0.5$. The results of the separations are shown in Figure \ref{fig:rectangle3d_manifolds}. The two groups of eigenvectors clearly show cosine waves running alongside the $x$ and $y$ directions, which precisely matches the actual eigenfunctions for this domain.

\subsection{Analysis of gateway criteria}

\begin{figure*}[ht]
    \centering
    \includegraphics[width=\linewidth]{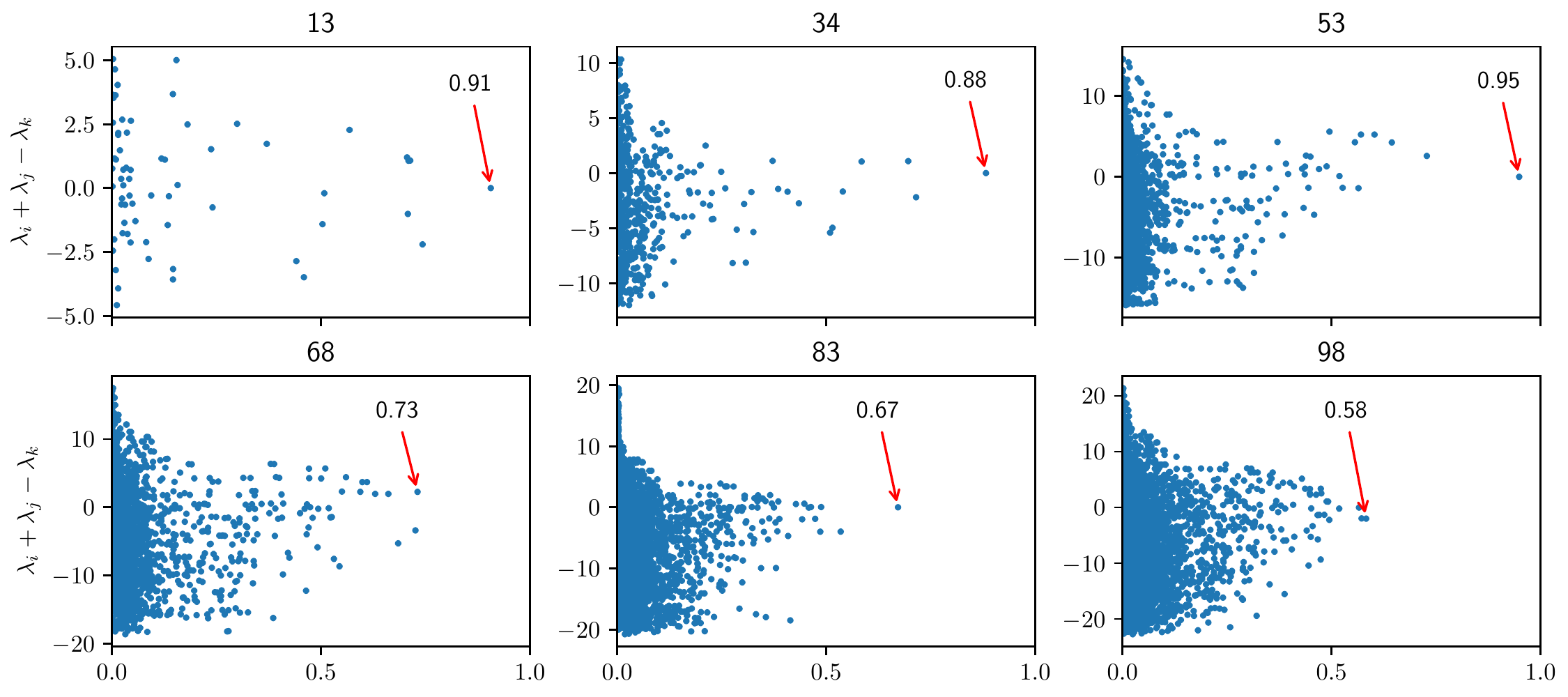}
    \caption{\textbf{Analysis of eigenvalue criterion and similarity score for the rectangle in 3D.} We choose six eigenvectors uniformly from the set of product eigenvectors and plot the eigenvalue criteria error and similarity scores for all possible triplets. Indices at the top identify the product eigenvector to which a column corresponds. The highest similarity is marked by a red arrow. The similarity is measured along the $x$-axis from 0 to 1.}
\label{fig:criterion-analysis}
\end{figure*}

Our algorithm has two main input parameters which act as thresholds: $\delta$ for the eigenvalue criterion and $\gamma$ for the similarity criterion. A question one might have is, for a given product eigenvector $\varphi_k$, how much of an outlier will the similarity of the true combination of factor eigenvectors $\varphi_i \varphi_j$ be compared to the similarities of all other pairs? To investigate this, we create scatterplots of the similarities against the eigenvalue criteria error over all pairs for various product eigenvectors, shown in Figure~\ref{fig:criterion-analysis}. We see that in most cases the similarity score for the chosen triplet is indeed an outlier from the rest of the distribution. Unsurprisingly, for product eigenvectors with lower eigenvalues the highest similarity stands out more significantly, as this part of the spectrum is more stable. As the eigenvalue of the product eigenvector increases, the best pair, or the one that has the highest similarity score and satisfies the eigenvalue criterion, becomes slightly more ambiguous. However, this ambiguity tends to be between relatively few candidates. This confirms that the eigenvalue and similarity criteria work well in conjunction.

\subsection{Single-particle cryo-EM data}

A particularly compelling application of manifold factorization is for cryo-electron microscopy (cryo-EM). Cryo-EM is a technique used for imaging proteins and other macromolecules that has fostered major advances in structural biology \citep{Kuhlbrandt2014}, thus garnering it recognition by the 2017 Nobel Prize in Chemistry \citep{CryoEMNobel2017}. More recently, cryo-EM has played a significant role in unraveling the structural properties of the novel coronavirus SARS-CoV-2 \citep{Zimmer2020}. 

In cryo-EM, molecular samples are rapidly frozen in a thin layer of ice, thus capturing them in their native  states \citep{Dubochet1988}. The preservation of this natural state is  key for better understanding of their biological function. Once frozen, a transmission electron microscope is used to produce images of the molecules, which are 2D tomographic projections of their electrostatic potential \citep{Vulovic2013}. These 2D images are then used to reconstruct, through a complicated series of computational steps, a 3D reconstruction of the molecule. 

The classical cryo-EM reconstruction problem is posed as follows: given $n$ tomographic projection images $I_1, \ldots, I_n$ of a particular molecule at random (unknown) orientations, how can we recover the three-dimensional structure of the molecule? A core obstacle to this problem is handling the low signal-to-noise ratio present in the micrographs \citep{SingerSigworth2020}. This task is made even more difficult if we consider the \textit{heterogeneity}  problem, which has been the subject of recent literature \citep{Frank2018,NakaneEtal2018,AndenSinger2018,SorzanoEtal2019,LedermanAndenSinger2020,ZhongBeplerDavisBerger2020}.
Here, a separate challenge arises from  structural variations arising from motion within the molecule. Namely, for different
images $I_i$ we will likely capture different molecular conformations.
In particular, several works have focused on the application of diffusion maps for the analysis of cryo-EM datasets with continuous hetetogeneity \citep{DashtiEtal2014,SchwanderFungOurmazd2014,MoscovichHaleviAndenSinger2020,ZeleskoMoscovichKileelSinger2020,DashtiEtal2020}.

For molecules that exhibit two or more independent continuous motions, the molecular shape manifold of 3D electrostatic densities is a product space, so one can try using our algorithm to factor it. 
For our experiments we used the model of a potassium ion channel
% Note: re-insert reference to Moscovich2019 for potassium ion channel model after AISTATS anonymized review
with two independently deformable parts: a spinning subunit with 360$^\circ$ of rotation, and a subunit that stretches in the $xy$-plane (see Figure~\ref{fig:cryo_x-theta_noisy_manifolds}). To create the images, we sampled $10,000$ conformations of the molecule with uniformly random angles and $x,y$ stretches ranging from $[-20, 20]$ in either direction. All the images were projected from a single view.
This models the computational pipeline of \citep{DashtiEtal2020}, which computes a diffusion maps embedding for each viewing direction separately. % Note: re-insert the reference and link to the ASPIRE repository after AISTATS anonymized review
Finally, we use PCA with four components to preprocess the raw images and transform them so that they are zero-centered with unit variance.

We perform three different trials: first, a dataset of molecules with rotations and stretches in the $x$ direction only and low noise; then, a dataset of molecules with rotations and stretches in the $x$-direction only and high noise; and finally, a dataset of molecules with rotations and stretches in both the $x$ and $y$ directions and low noise. For each trial, use $\gamma = 0.80$ and $\delta = 1.0$. The results for the second trial are shown in Figure \ref{fig:cryo_x-theta_noisy_manifolds}. The eigenvectors corresponding to the $x$-stretch manifold exhibit the ground truth cosine waves, and the eigenvectors corresponding to the rotation manifold (a circle) exhibit the ground truth sine waves.
Note that even though the latent space is a product space, Assumption~\ref{assump:isometric} does not strictly hold due to the additive Gaussian noise.
Nonetheless, our method works nicely on this dataset.

\begin{figure}[t]
    \centering
    \def\arraystretch{1.5}
    \begin{tabular}{cc}
        \includegraphics[width=0.45\linewidth]{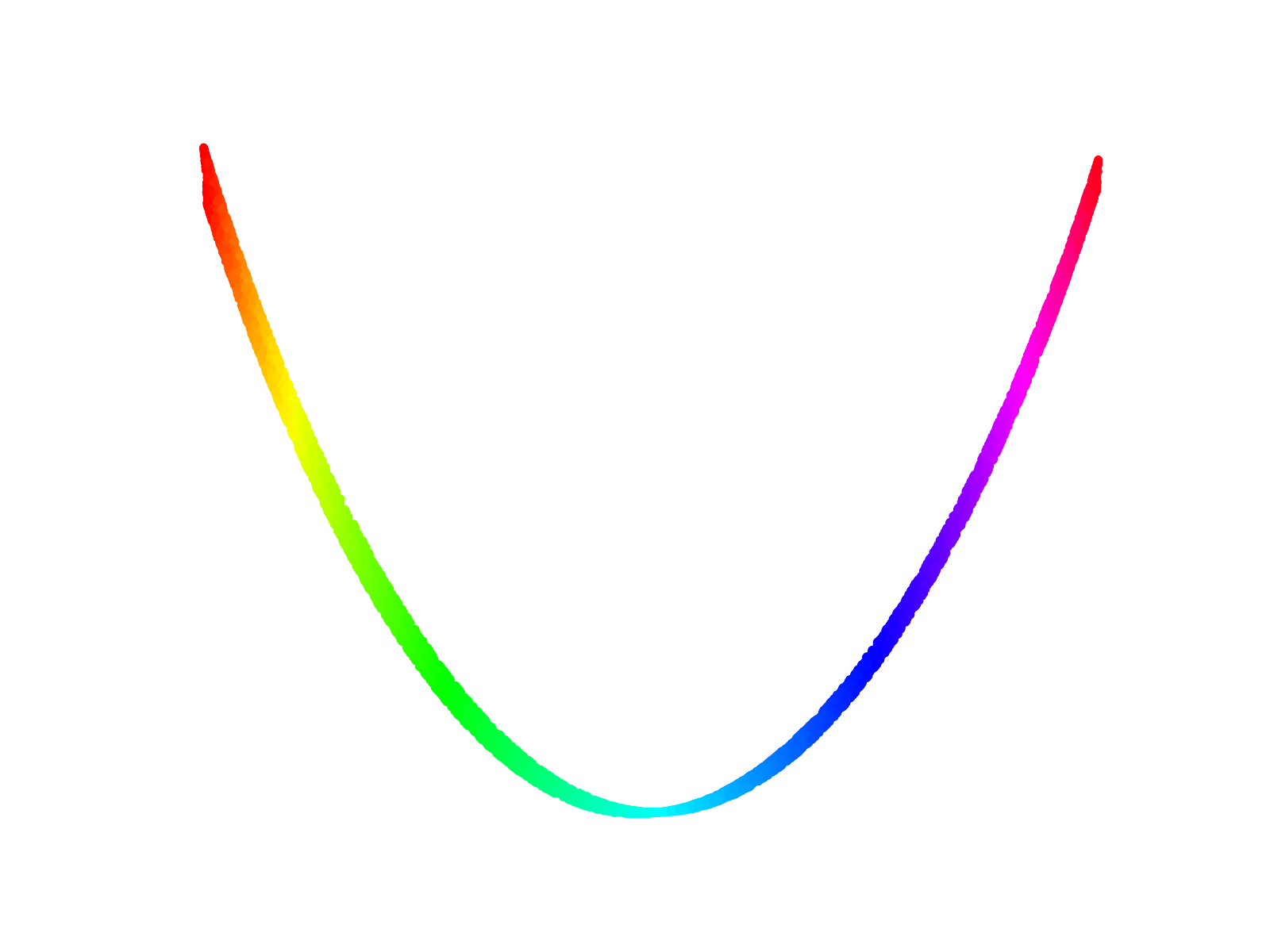} & 
        \includegraphics[width=0.45\linewidth]{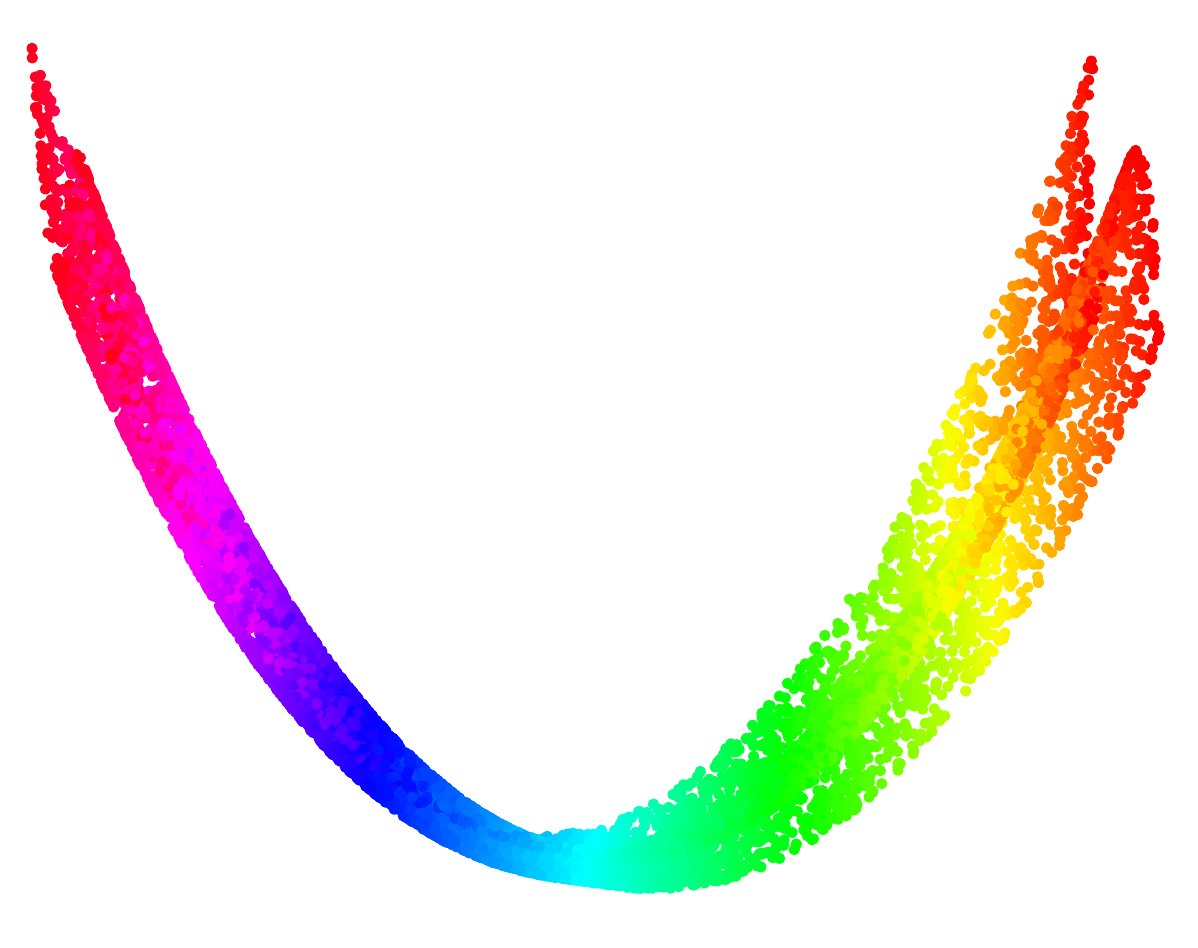} \\
        \includegraphics[width=0.45\linewidth]{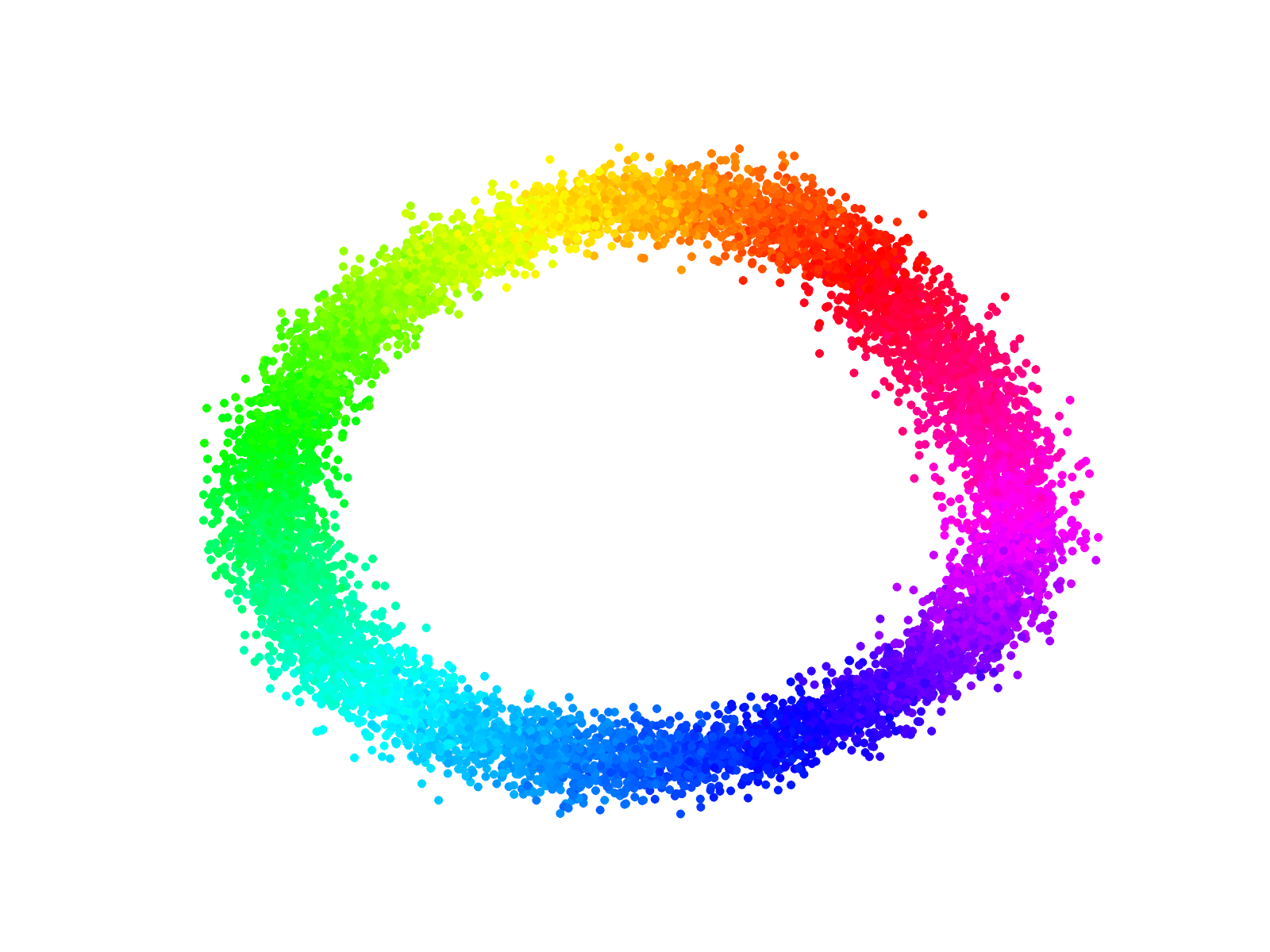} &     
        \includegraphics[width=0.45\linewidth]{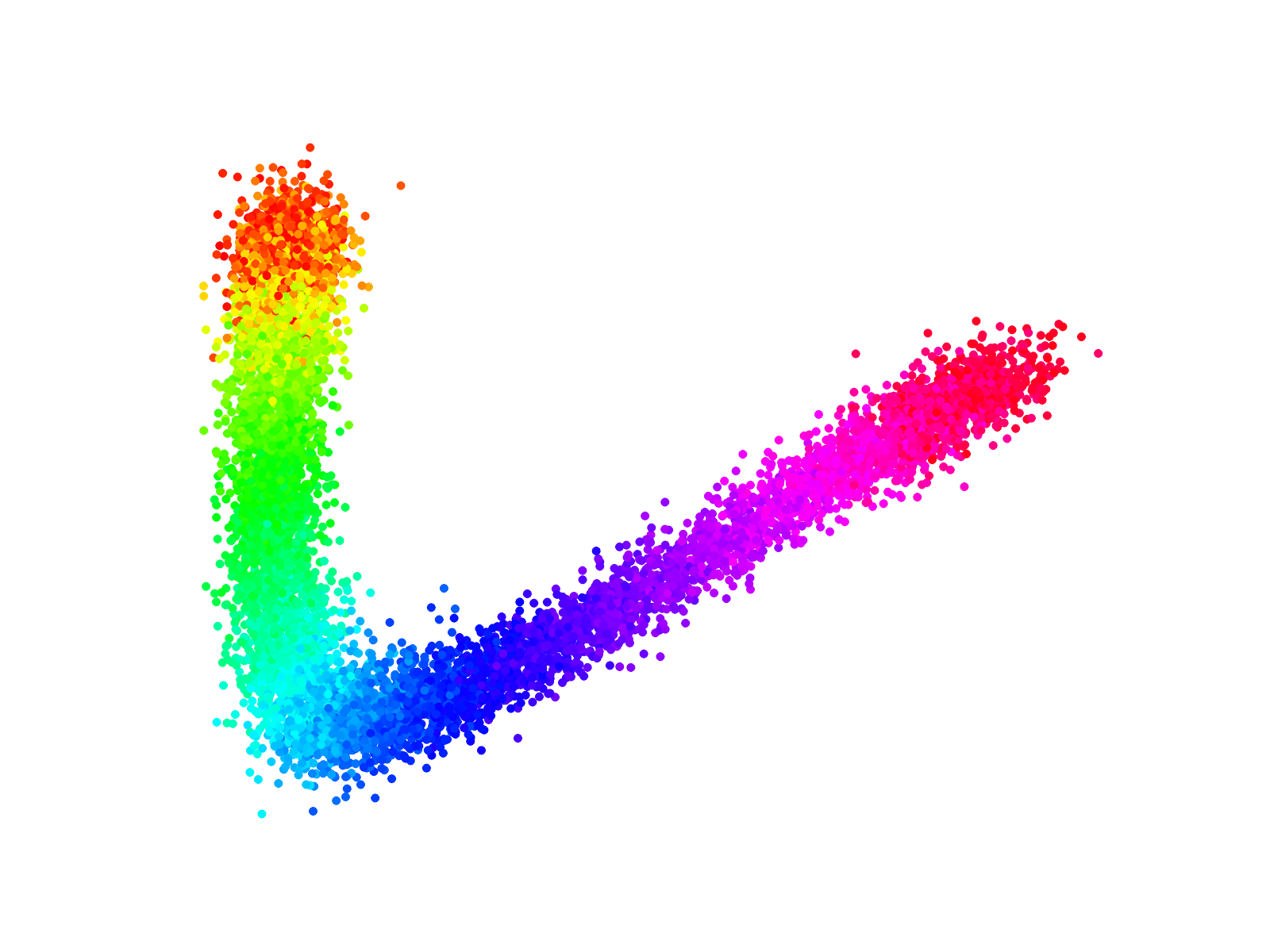} \\
        \includegraphics[width=0.45\linewidth]{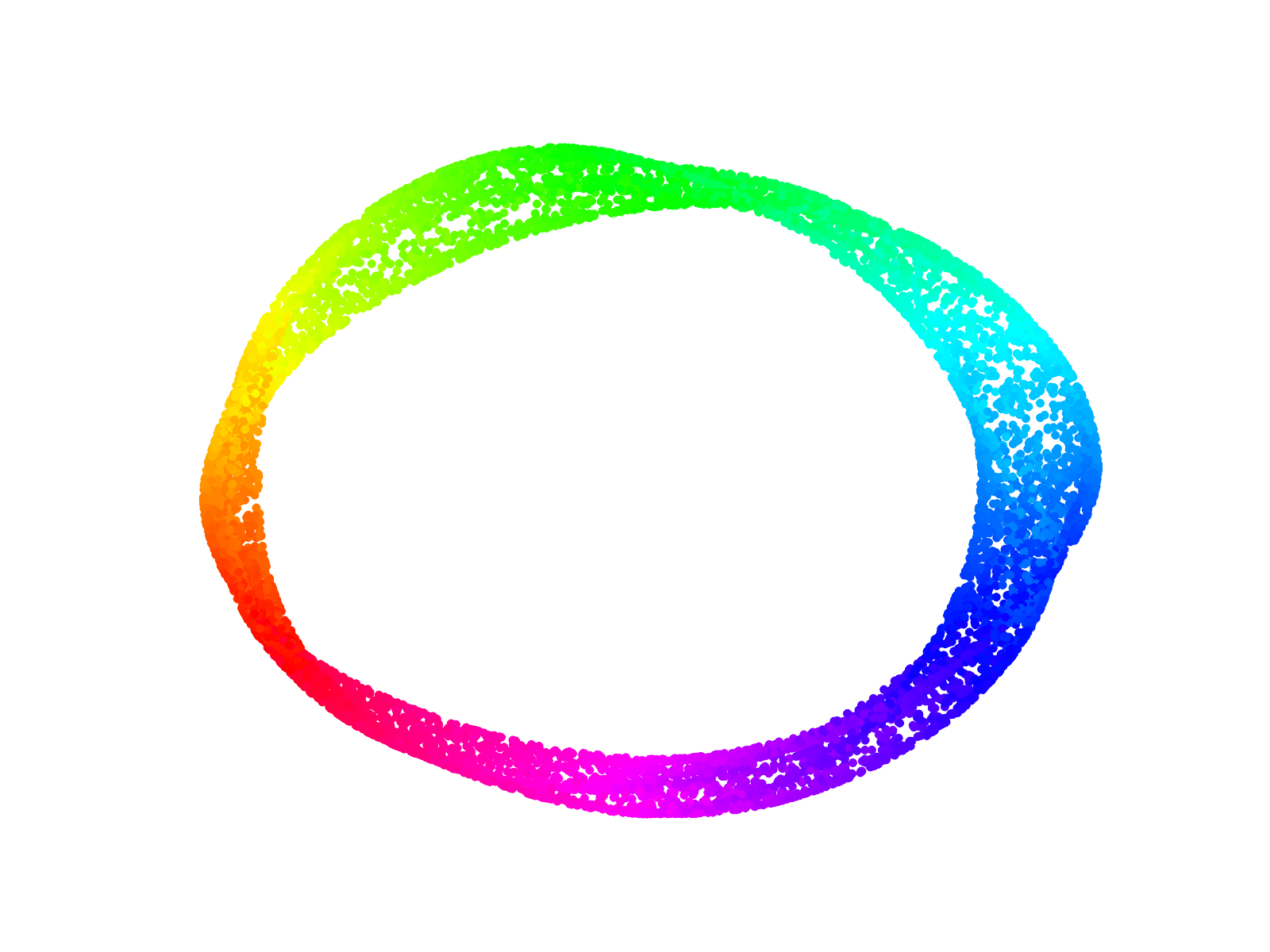} &     
        \includegraphics[width=0.45\linewidth]{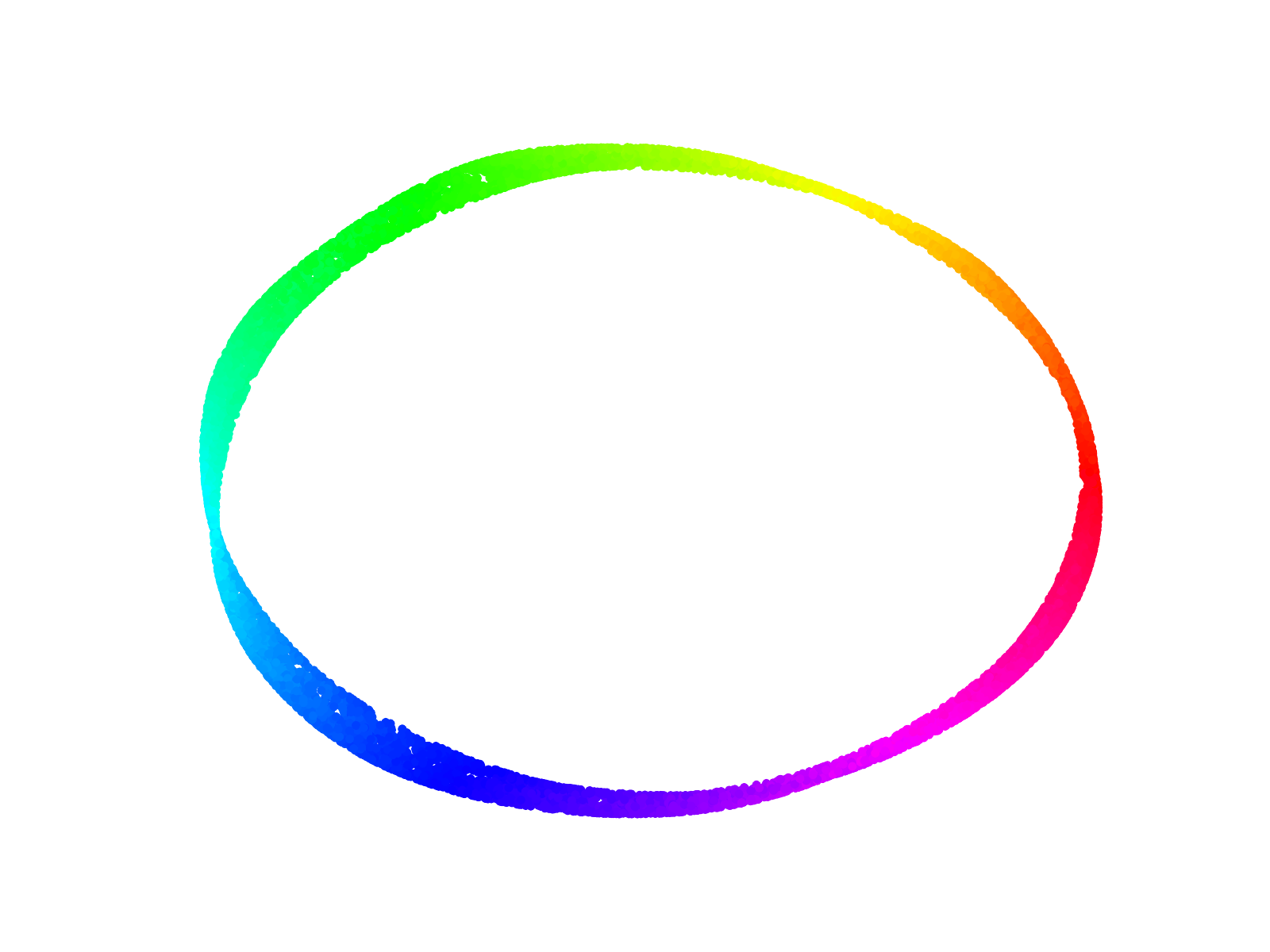} \\
        $\mathcal M_1$ & $\mathcal M_2$
    \end{tabular}
    \caption{\textbf{Laplacian eigenmap embeddings.} Each scatterplot shows 10,000 samples of the first two non-trivial eigenvectors found by the algorithm. The colors represent the ground truth data in the parameter space. \textbf{Top row.} Samples from a rectangle with $z$-direction noise. The algorithm was run with $\delta = 0.5$, $\gamma = 0.85$. \textbf{Middle row.} Samples from noisy cryo-EM data. The algorithm was run with $\delta = 1.0$, $\gamma = 0.85$. \textbf{Bottom row.} Samples from a torus. The algorithm was run with $\delta = 1.0$ and $\gamma = 0.6$.}
    \label{fig:eigenmap_embeddings}
\end{figure}

\subsection{Laplacian eigenmap embeddings}

We can visualize the manifold factors by plotting the Laplacian eigenmap embeddings. Figure \ref{fig:eigenmap_embeddings} shows the eigenmap embeddings for three different experiments: the rectangle with noise in the $z$-direction; noisy cryo-EM data with a rotational component and  a stretch component in the $x$-direction; and a torus. In each case, the embeddings are in agreement with the actual structure of the underlying domain.

These embeddings show that the results of the algorithm can be used to accomplish two objectives. First, we can use representative factor eigenvectors from each manifold factor to get a more condensed representation of the data, thus achieving dimensionality reduction. This is similar to using diffusion coordinates for dimensionality reduction, but diffusion coordinates generally appear in no particular order. Additionally, by separating the factor eigenvectors according to the separate manifold factors, we also ``de-mix" these diffusion coordinates into a more interpretable order.

\section{Conclusion}\label{conclusion}

In this paper, we presented a spectral method for non-linear dimensionality reduction and data representation.
Our method is based on manifold factorization and applies to data sets with two or more independent degrees of freedom.
We tested our algorithm on electron-microscope snapshots of a protein with two independently moving subunits and demonstrated that we can recover the motion manifolds for each subunit.
Hence, our approach shows promise as a tool for the analysis of macromolecules with continuous degrees of freedom.

While our discussion and experiments are limited to the two-factor case, our algorithm can be extended to the general case of $K$ independent latent variables by replacing \textsc{Max-Cut} with \textsc{Max-K-Cut} \citep{Newman2018} or similar procedures.
Another interesting direction for future work is to study the potential of our approach for analyzing broader classes of data sets, other than product spaces, where the Laplacian operator is still separable.

\subsection*{Reproducibility}
Code for reproducing the figures in this paper can be found at: \url{https://github.com/sxzhang25/product-manifold-learning}

\addcontentsline{toc}{section}{References}
\bibliography{product-manifold-learning}

\begin{thebibliography}{53}
\providecommand{\natexlab}[1]{#1}
\providecommand{\url}[1]{\texttt{#1}}
\expandafter\ifx\csname urlstyle\endcsname\relax
  \providecommand{\doi}[1]{doi: #1}\else
  \providecommand{\doi}{doi: \begingroup \urlstyle{rm}\Url}\fi

\bibitem[And{\'{e}}n and Singer(2018)]{AndenSinger2018}
Joakim And{\'{e}}n and Amit Singer.
\newblock {Structural Variability from Noisy Tomographic Projections}.
\newblock \emph{SIAM Journal on Imaging Sciences}, 11\penalty0 (2):\penalty0
  1441--1492, 2018.
\newblock \doi{10.1137/17M1153509}.

\bibitem[Bates(2014)]{Bates2014}
Jonathan Bates.
\newblock {The embedding dimension of Laplacian eigenfunction maps}.
\newblock \emph{Applied and Computational Harmonic Analysis}, 37\penalty0
  (3):\penalty0 516--530, 2014.
\newblock \doi{10.1016/j.acha.2014.03.002}.

\bibitem[Belkin and Niyogi(2003)]{BelkinNiyogi2003}
Mikhail Belkin and Partha Niyogi.
\newblock {Laplacian Eigenmaps for Dimensionality Reduction and Data
  Representation}.
\newblock \emph{Neural Computation}, 15\penalty0 (6):\penalty0 1373--1396,
  2003.
\newblock \doi{10.1162/089976603321780317}.

\bibitem[Belkin and Niyogi(2008)]{BelkinNiyogi2008}
Mikhail Belkin and Partha Niyogi.
\newblock {Towards a theoretical foundation for Laplacian-based manifold
  methods}.
\newblock \emph{Journal of Computer and System Sciences}, 74\penalty0
  (8):\penalty0 1289--1308, 2008.
\newblock \doi{10.1016/j.jcss.2007.08.006}.

\bibitem[Canzani(2013)]{Canzani2013}
Yaiza Canzani.
\newblock {Analysis on Manifolds via the Laplacian (lecture notes)}, 2013.

\bibitem[Coifman and Lafon(2006)]{CoifmanLafon2006}
Ronald~R. Coifman and St{\'{e}}phane Lafon.
\newblock {Diffusion maps}.
\newblock \emph{Applied and Computational Harmonic Analysis}, 21\penalty0
  (1):\penalty0 5--30, 2006.
\newblock \doi{10.1016/j.acha.2006.04.006}.

\bibitem[Cressey and Callaway(2017)]{CryoEMNobel2017}
Daniel Cressey and Ewen Callaway.
\newblock {Cryo-electron microscopy wins chemistry Nobel}.
\newblock \emph{Nature}, 550\penalty0 (7675):\penalty0 167--167, 2017.
\newblock \doi{10.1038/nature.2017.22738}.

\bibitem[Dashti et~al.(2014)Dashti, Schwander, Langlois, Fung, Li,
  Hosseinizadeh, Liao, Pallesen, Sharma, Stupina, Simon, Dinman, Frank, and
  Ourmazd]{DashtiEtal2014}
Ali Dashti et~al.
\newblock {Trajectories of the ribosome as a Brownian nanomachine}.
\newblock \emph{Proceedings of the National Academy of Sciences}, 111\penalty0
  (49):\penalty0 17492--17497, 2014.
\newblock \doi{10.1073/pnas.1419276111}.

\bibitem[Dashti et~al.(2020)Dashti, Mashayekhi, Shekhar, {Ben Hail}, Salah,
  Schwander, des Georges, Singharoy, Frank, and Ourmazd]{DashtiEtal2020}
Ali Dashti et~al.
\newblock {Retrieving functional pathways of biomolecules from single-particle
  snapshots}.
\newblock \emph{Nature Communications}, 11\penalty0 (1):\penalty0 4734, 2020.
\newblock \doi{10.1038/s41467-020-18403-x}.

\bibitem[Diamond and Boyd(2016)]{siamond2016cvxpy}
Steven Diamond and Stephen Boyd.
\newblock {CVXPY: A Python-Embedded Modeling Language for Convex Optimization}.
\newblock Technical report, 2016.
\newblock \url{http://www.cvxpy.org/}.

\bibitem[Dubochet et~al.(1988)Dubochet, Adrian, Chang, Homo, Lepault, McDowall,
  and Schultz]{Dubochet1988}
Jacques Dubochet et~al.
\newblock {Cryo-electron microscopy of vitrified specimens}.
\newblock \emph{Quarterly Reviews of Biophysics}, 21\penalty0 (2):\penalty0
  129--228, 1988.
\newblock \doi{10.1017/S0033583500004297}.

\bibitem[Folland(1976)]{Folland1976}
Gerald~B. Folland.
\newblock \emph{{Introduction to Partial Differential Equations}}.
\newblock Princeton University Press, 1976.
\newblock ISBN 9780691213033.
\newblock \doi{10.2307/j.ctvzsmfgn}.

\bibitem[Frank(2018)]{Frank2018}
Joachim Frank.
\newblock {New Opportunities Created by Single-Particle Cryo-EM: The Mapping of
  Conformational Space}.
\newblock \emph{Biochemistry}, 57\penalty0 (6):\penalty0 888--888, 2018.
\newblock \doi{10.1021/acs.biochem.8b00064}.

\bibitem[{Garc{\'{i}}a Trillos} and Slep{\v{c}}ev(2018)]{TrillosSlepcev2018}
Nicol{\'{a}}s {Garc{\'{i}}a Trillos} and Dejan Slep{\v{c}}ev.
\newblock {A variational approach to the consistency of spectral clustering}.
\newblock \emph{Applied and Computational Harmonic Analysis}, 45\penalty0
  (2):\penalty0 239--281, 2018.
\newblock \doi{10.1016/j.acha.2016.09.003}.

\bibitem[{Garc{\'{i}}a Trillos} et~al.(2020){Garc{\'{i}}a Trillos}, Gerlach,
  Hein, and Slep{\v{c}}ev]{TrillosGerlachHeinSlepcev2020}
Nicol{\'{a}}s {Garc{\'{i}}a Trillos}, Moritz Gerlach, Matthias Hein and Dejan
  Slep{\v{c}}ev.
\newblock {Error Estimates for Spectral Convergence of the Graph Laplacian on
  Random Geometric Graphs Toward the {Laplace--Beltrami} Operator}.
\newblock \emph{Foundations of Computational Mathematics}, 20\penalty0
  (4):\penalty0 827--887, 2020.
\newblock \doi{10.1007/s10208-019-09436-w}.

\bibitem[Goemans and Williamson(1995)]{Goemans1995}
Michel~X Goemans and David~P Williamson.
\newblock {Improved approximation algorithms for maximum cut and satisfiability
  problems using semidefinite programming}.
\newblock \emph{Journal of the ACM}, 42\penalty0 (6):\penalty0 1115--1145,
  1995.
\newblock \doi{10.1145/227683.227684}.

\bibitem[Grebenkov and Nguyen(2013)]{GrebenkovNguyen2013}
Denis~S. Grebenkov and Binh-Thanh Nguyen.
\newblock {Geometrical Structure of Laplacian Eigenfunctions}.
\newblock \emph{SIAM Review}, 55\penalty0 (4):\penalty0 601--667, 2013.
\newblock \doi{10.1137/120880173}.

\bibitem[Hein et~al.(2005)Hein, Audibert, and von
  Luxburg]{HeinAudibertLuxburg2005}
Matthias Hein, Jean-Yves Audibert and Ulrike von Luxburg.
\newblock {From Graphs to Manifolds – Weak and Strong Pointwise Consistency
  of Graph Laplacians}.
\newblock In \emph{International Conference on Computational Learning Theory
  (COLT)}, pages 470--485, 2005.
\newblock \doi{10.1007/11503415_32}.

\bibitem[Kim and Mnih(2018)]{Kim2018}
Hyunjik Kim and Andriy Mnih.
\newblock {Disentangling by factorising}.
\newblock In \emph{International Conference on Machine Learning (ICML)},
  volume~6, pages 4153--4171, 2018.

\bibitem[K{\"u}hlbrandt(2014)]{Kuhlbrandt2014}
W.~K{\"u}hlbrandt.
\newblock {The Resolution Revolution}.
\newblock \emph{Science}, 343\penalty0 (6178):\penalty0 1443--1444, 2014.
\newblock \doi{10.1126/science.1251652}.

\bibitem[Kuiper(1955)]{Kuiper1955}
Nicolaas~H. Kuiper.
\newblock {On C1-isometric imbeddings. I}.
\newblock \emph{Indagationes Mathematicae (Proceedings)}, 58:\penalty0
  545--556, 1955.
\newblock \doi{10.1016/S1385-7258(55)50075-8}.

\bibitem[Lafon(2004)]{Lafon2004}
Stephane~S. Lafon.
\newblock \emph{{Diffusion Maps and Geometric Harmonics}}.
\newblock PhD thesis, Yale University, 2004.

\bibitem[Lederman et~al.(2020)Lederman, And{\'{e}}n, and
  Singer]{LedermanAndenSinger2020}
Roy~R. Lederman, Joakim And{\'{e}}n and Amit Singer.
\newblock {Hyper-molecules: on the representation and recovery of dynamical
  structures for applications in flexible macro-molecules in cryo-EM}.
\newblock \emph{Inverse Problems}, 36\penalty0 (4):\penalty0 044005, 2020.
\newblock \doi{10.1088/1361-6420/ab5ede}.

\bibitem[Lee and Izbicki(2016)]{LeeIzbicki2016}
Ann~B. Lee and Rafael Izbicki.
\newblock {A spectral series approach to high-dimensional nonparametric
  regression}.
\newblock \emph{Electronic Journal of Statistics}, 10\penalty0 (1):\penalty0
  423--463, 2016.
\newblock \doi{10.1214/16-EJS1112}.

\bibitem[Lee(2012)]{Lee2012}
John~M. Lee.
\newblock \emph{{Introduction to Smooth Manifolds}}, volume 218 of
  \emph{Graduate Texts in Mathematics}.
\newblock Springer New York, 2012.
\newblock ISBN 978-1-4419-9981-8.
\newblock \doi{10.1007/978-1-4419-9982-5}.

\bibitem[Locatello et~al.(2019)Locatello, Bauer, Lucic, R{\"a}tsch, Gelly,
  Sch{\"o}lkopf, and Bachem]{LocatelloEtal2019}
Francesco Locatello et~al.
\newblock {Challenging Common Assumptions in the Unsupervised Learning of
  Disentangled Representations}.
\newblock \emph{International Conference on Machine Learning (ICML)}, pages
  4114--4124, 2019.

\bibitem[Martinsson et~al.(2011)Martinsson, Rokhlin, and
  Tygert]{Martinsson2011}
Per~Gunnar Martinsson, Vladimir Rokhlin and Mark Tygert.
\newblock {A randomized algorithm for the decomposition of matrices}.
\newblock \emph{Applied and Computational Harmonic Analysis}, 30\penalty0
  (1):\penalty0 47--68, 2011.
\newblock \doi{10.1016/j.acha.2010.02.003}.

\bibitem[Moscovich et~al.(2020)Moscovich, Halevi, And{\'{e}}n, and
  Singer]{MoscovichHaleviAndenSinger2020}
Amit Moscovich, Amit Halevi, Joakim And{\'{e}}n and Amit Singer.
\newblock {Cryo-EM reconstruction of continuous heterogeneity by Laplacian
  spectral volumes}.
\newblock \emph{Inverse Problems}, 36\penalty0 (2):\penalty0 024003, 2020.
\newblock \doi{10.1088/1361-6420/ab4f55}.

\bibitem[Nadler et~al.(2005)Nadler, Lafon, Coifman, and
  Kevrekidis]{NadlerLafonCoifmanKevrekidis2005}
Boaz Nadler, Stephane Lafon, Ronald~R. Coifman and Ioannis~G. Kevrekidis.
\newblock {Diffusion Maps, Spectral Clustering and Eigenfunctions of
  Fokker-Planck operators}.
\newblock In \emph{Advances in Neural Information Processing Systems (NIPS)},
  pages 955--962, 2005.

\bibitem[Nakane et~al.(2018)Nakane, Kimanius, Lindahl, and
  Scheres]{NakaneEtal2018}
Takanori Nakane, Dari Kimanius, Erik Lindahl and Sjors~HW Scheres.
\newblock {Characterisation of molecular motions in cryo-EM single-particle
  data by multi-body refinement in RELION}.
\newblock \emph{eLife}, 7:\penalty0 1--18, 2018.
\newblock \doi{10.7554/eLife.36861}.

\bibitem[Nash(1954)]{Nash1954}
John Nash.
\newblock {{\(C^1\)} Isometric Imbeddings}.
\newblock \emph{The Annals of Mathematics}, 60\penalty0 (3):\penalty0 383--396,
  1954.
\newblock \doi{10.2307/1969840}.

\bibitem[Newman(2018)]{Newman2018}
Alantha Newman.
\newblock {Complex semidefinite programming and max-k-cut}.
\newblock In \emph{Symposium on Simplicity in Algorithms (SOSA)}, volume~61,
  pages 13:1--13:11, 2018.
\newblock \doi{10.4230/OASIcs.SOSA.2018.13}.

\bibitem[Pedregosa et~al.(2011)Pedregosa, Varoquaux, Gramfort, Michel, Thirion,
  Grisel, Blondel, Prettenhofer, Weiss, Dubourg, Vanderplas, Passos,
  Cournapeau, Brucher, Perrot, and Duchesnay]{scikitlearn2011}
Fabian Pedregosa et~al.
\newblock {Scikit-learn: Machine Learning in {P}ython}.
\newblock \emph{Journal of Machine Learning Research}, 12:\penalty0 2825--2830,
  2011.

\bibitem[Reed et~al.(2014)Reed, Sohn, Zhang, and Lee]{ReedSohnZhangLee2014}
Scott Reed, Kihyuk Sohn, Yuting Zhang and Honglak Lee.
\newblock {Learning to disentangle factors of variation with manifold
  interaction}.
\newblock \emph{International Conference on Machine Learning (ICML)}, pages
  1431--1439, 2014.

\bibitem[Rodol{\`{a}} et~al.(2019)Rodol{\`{a}}, L{\"{a}}hner, Bronstein,
  Bronstein, and Solomon]{Rodola2019}
Emanuele Rodol{\`{a}}, Zorah L{\"{a}}hner, Alex~M. Bronstein, Michael~M.
  Bronstein and Justin Solomon.
\newblock {Functional Maps Representation On Product Manifolds}.
\newblock \emph{Computer Graphics Forum}, 38\penalty0 (1):\penalty0 678--689,
  2019.
\newblock \doi{10.1111/cgf.13598}.

\bibitem[Rosasco et~al.(2010)Rosasco, Belkin, and {De
  Vito}]{RosascoBelkinDevito2010}
Lorenzo Rosasco, Mikhail Belkin and Ernesto {De Vito}.
\newblock {On Learning with Integral Operators}.
\newblock \emph{Journal of Machine Learning Research}, 11:\penalty0 905--934,
  2010.

\bibitem[Schwander et~al.(2014)Schwander, Fung, and
  Ourmazd]{SchwanderFungOurmazd2014}
Peter Schwander, Russell Fung and Abbas Ourmazd.
\newblock {Conformations of macromolecules and their complexes from
  heterogeneous datasets}.
\newblock \emph{Philosophical Transactions of the Royal Society B: Biological
  Sciences}, 369\penalty0 (1647):\penalty0 1--8, 2014.
\newblock \doi{10.1098/rstb.2013.0567}.

\bibitem[Siddharth et~al.(2017)Siddharth, Paige, {Van De Meent}, Desmaison,
  Goodman, Kohli, Wood, and Torr]{Siddharth2017}
N~Siddharth et~al.
\newblock {Learning Disentangled Representations with Semi-Supervised Deep
  Generative Models}.
\newblock In \emph{Conference on Neural Information Processing Systems (NIPS)},
  2017.

\bibitem[Singer(2006{\natexlab{a}})]{Singer2006a}
A.~Singer.
\newblock {From graph to manifold Laplacian: The convergence rate}.
\newblock \emph{Applied and Computational Harmonic Analysis}, 21\penalty0
  (1):\penalty0 128--134, 2006{\natexlab{a}}.
\newblock \doi{10.1016/j.acha.2006.03.004}.

\bibitem[Singer(2006{\natexlab{b}})]{Singer2006b}
A.~Singer.
\newblock {Spectral independent component analysis}.
\newblock \emph{Applied and Computational Harmonic Analysis}, 21\penalty0
  (1):\penalty0 135--144, 2006{\natexlab{b}}.
\newblock \doi{10.1016/j.acha.2006.03.003}.

\bibitem[Singer and Coifman(2008)]{SingerCoifman2008}
Amit Singer and Ronald~R. Coifman.
\newblock {Non-linear independent component analysis with diffusion maps}.
\newblock \emph{Applied and Computational Harmonic Analysis}, 25\penalty0
  (2):\penalty0 226--239, 2008.
\newblock \doi{10.1016/j.acha.2007.11.001}.

\bibitem[Singer and Sigworth(2020)]{SingerSigworth2020}
Amit Singer and Fred~J. Sigworth.
\newblock {Computational Methods for Single-Particle Electron Cryomicroscopy}.
\newblock \emph{Annual Review of Biomedical Data Science}, 3\penalty0
  (1):\penalty0 163--190, 2020.
\newblock \doi{10.1146/annurev-biodatasci-021020-093826}.

\bibitem[Sorzano et~al.(2019)Sorzano, Jim{\'{e}}nez, Mota, Vilas, Maluenda,
  Mart{\'{i}}nez, Ram{\'{i}}rez-Aportela, Majtner, Segura,
  S{\'{a}}nchez-Garc{\'{i}}a, Rancel, del Ca{\~{n}}o, Conesa, Melero, Jonic,
  Vargas, Cazals, Freyberg, Krieger, Bahar, Marabini, and
  Carazo]{SorzanoEtal2019}
Carlos Oscar~S. Sorzano et~al.
\newblock {Survey of the analysis of continuous conformational variability of
  biological macromolecules by electron microscopy}.
\newblock \emph{Acta Crystallographica Section F Structural Biology
  Communications}, 75\penalty0 (1):\penalty0 19--32, 2019.
\newblock \doi{10.1107/S2053230X18015108}.

\bibitem[Talmon et~al.(2013)Talmon, Cohen, Gannot, and
  Coifman]{TalmonCohenGannotCoifman2013}
Ronen Talmon, Israel Cohen, Sharon Gannot and Ronald~R. Coifman.
\newblock {Diffusion Maps for Signal Processing: A Deeper Look at
  Manifold-Learning Techniques Based on Kernels and Graphs}.
\newblock \emph{IEEE Signal Processing Magazine}, 30\penalty0 (4):\penalty0
  75--86, 2013.
\newblock \doi{10.1109/MSP.2013.2250353}.

\bibitem[Tenenbaum et~al.(2000)Tenenbaum, de~Silva, and
  Langford]{TenenbaumDesilvaLangford2000}
Joshua~B. Tenenbaum, Vin de~Silva and John~C. Langford.
\newblock {A Global Geometric Framework for Nonlinear Dimensionality
  Reduction}.
\newblock \emph{Science}, 290\penalty0 (5500):\penalty0 2319--2323, 2000.
\newblock \doi{10.1126/science.290.5500.2319}.

\bibitem[Ting et~al.(2010)Ting, Huang, and Jordan]{TingHuangJordan2010}
Daniel Ting, Ling Huang and Michael Jordan.
\newblock {An Analysis of the Convergence of Graph Laplacians}.
\newblock In \emph{International Conference on Machine Learning (ICML)}, 2010.

\bibitem[Tran et~al.(2017)Tran, Yin, and Liu]{Tran2017}
Luan Tran, Xi~Yin and Xiaoming Liu.
\newblock {Disentangled Representation Learning GAN for Pose-Invariant Face
  Recognition}.
\newblock In \emph{Conference on Computer Vision and Pattern Recognition
  (CVPR)}, pages 1283--1292, 2017.
\newblock \doi{10.1109/CVPR.2017.141}.

\bibitem[{Von Luxburg}(2007)]{VonLuxburg2007}
Ulrike {Von Luxburg}.
\newblock {A tutorial on spectral clustering}.
\newblock \emph{Statistics and Computing}, 17\penalty0 (4):\penalty0 395--416,
  2007.
\newblock \doi{10.1007/s11222-007-9033-z}.

\bibitem[von Luxburg et~al.(2008)von Luxburg, Belkin, and
  Bousquet]{VonLuxburgBelkinBousquet2008}
Ulrike von Luxburg, Mikhail Belkin and Olivier Bousquet.
\newblock {Consistency of spectral clustering}.
\newblock \emph{The Annals of Statistics}, 36\penalty0 (2):\penalty0 555--586,
  2008.
\newblock \doi{10.1214/009053607000000640}.

\bibitem[Vulovi{\'{c}} et~al.(2013)Vulovi{\'{c}}, Ravelli, van Vliet, Koster,
  Lazi{\'{c}}, L{\"{u}}cken, Rullg{\aa}rd, {\"{O}}ktem, and
  Rieger]{Vulovic2013}
Milo{\v{s}} Vulovi{\'{c}} et~al.
\newblock {Image formation modeling in cryo-electron microscopy}.
\newblock \emph{Journal of Structural Biology}, 183\penalty0 (1):\penalty0
  19--32, 2013.
\newblock \doi{10.1016/j.jsb.2013.05.008}.

\bibitem[Zelesko et~al.(2020)Zelesko, Moscovich, Kileel, and
  Singer]{ZeleskoMoscovichKileelSinger2020}
Nathan Zelesko, Amit Moscovich, Joe Kileel and Amit Singer.
\newblock {Earthmover-Based Manifold Learning for Analyzing Molecular
  Conformation Spaces}.
\newblock In \emph{International Symposium on Biomedical Imaging (ISBI)}, pages
  1715--1719, 2020.
\newblock \doi{10.1109/ISBI45749.2020.9098723}.

\bibitem[Zhong et~al.(2020)Zhong, Bepler, Davis, and
  Berger]{ZhongBeplerDavisBerger2020}
Ellen~D. Zhong, Tristan Bepler, Joseph~H. Davis and Bonnie Berger.
\newblock {Reconstructing continuous distributions of 3D protein structure from
  cryo-{EM} images}.
\newblock In \emph{International Conference on Learning Representations
  (ICLR)}, pages 1--20, 2020.

\bibitem[Zimmer(2020)]{Zimmer2020}
Carl Zimmer.
\newblock {The Coronavirus Unveiled}, oct 2020, The New York Times.
\newblock \url{http://nyti.ms/2GO6PDV}.

\end{thebibliography}
\bibliographystyle{plainnat-edited}
\end{document}